%% file: sample-authordraft.tex
\documentclass[acmtog, nonacm]{acmart}

\AtBeginDocument{%
  \providecommand\BibTeX{{%
    \normalfont B\kern-0.5em{\scshape i\kern-0.25em b}\kern-0.8em\TeX}}}

\usepackage{algorithm2e}
\definecolor{myblue}{rgb}{0, 0, 0}
\usepackage{caption}

\begin{document}

\title{Spatially Adaptive Cloth Regression with Implicit Neural Representations}

\author{Lei Shu}
\email{leishu@student.ethz.ch}
\affiliation{%
  \institution{ETH Z\"urich \& Disney Research|Studios}
  \city{Z\"urich}
  \country{Switzerland}
}

\author{Vinicius Azevedo}
\email{vinicius.azevedo@disneyresearch.com}
\affiliation{%
  \institution{Disney Research|Studios}
  \city{Z\"urich}
  \country{Switzerland}
}

\author{Barbara Solenthaler}
\email{solenthaler@inf.ethz.ch}
\affiliation{%
  \institution{ETH Z\"urich}
  \city{Z\"urich}
  \country{Switzerland}
}

\author{Markus Gross}
\email{grossm@inf.ethz.ch}
\affiliation{%
  \institution{ETH Z\"urich \& Disney Research|Studios}
  \city{Z\"urich}
  \country{Switzerland}
}

\renewcommand{\shortauthors}{Shu, et al.}



\begin{teaserfigure}
  \includegraphics[width=\textwidth]{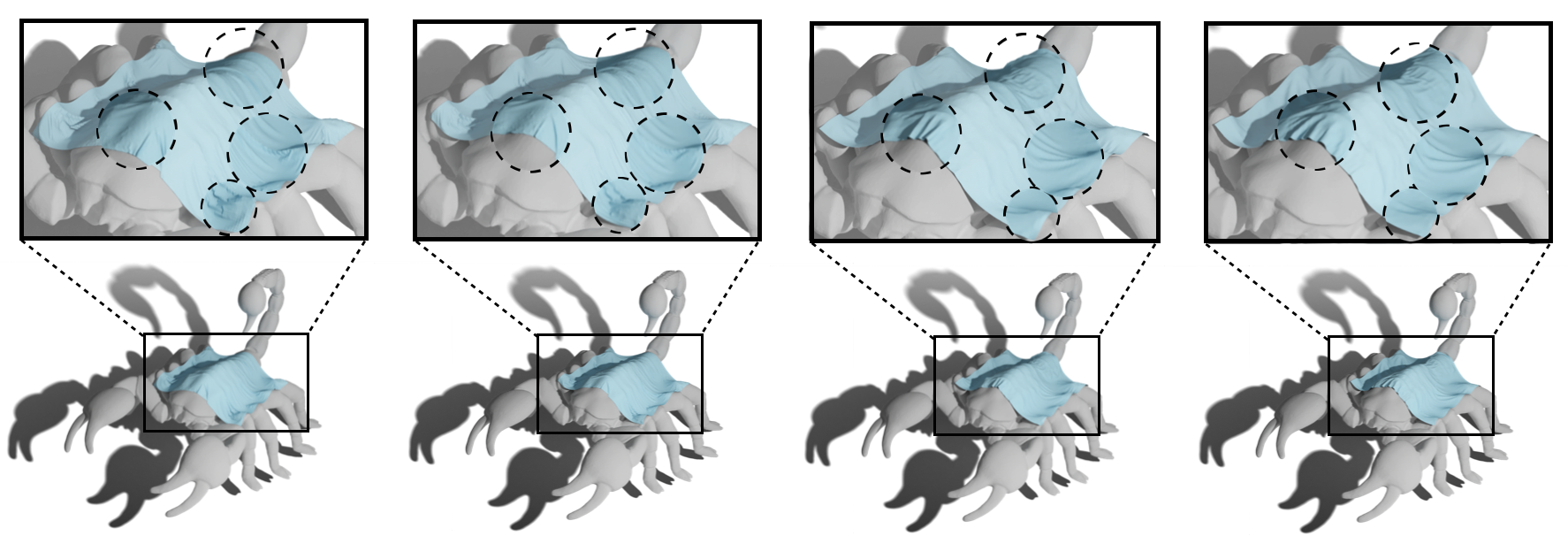}
  \caption{\textbf{Wrinkle Generation in Cloth-Object Interaction.} (Left) A coarse-resolution mesh grid (resolution \(128 \times 128\), totaling \(49152\) free variables) employs the original mesh connectivity for loss computation. Traditional meshes struggle to produce detailed wrinkles at low resolutions, and even generate unnatural artifacts due to discretization in some places. (Middle Left) A multi-resolution grid neural network with fewer free variables (\(47369\)) captures cloth details using the original mesh connectivity. This model shows small improvements over direct vertex optimization but still finds it challenging to capture detailed wrinkles correctly and naturally. (Middle Right) The same variables (\(47369\)) in the multi-resolution grid model, with losses computed using our novel method (Section ~\ref{sec:method}) but with \textit{uniform} sampling of \textit{local structures} (Subsection ~\ref{sec:3.3}). This continuous domain approach significantly enhances wrinkle patterns in a more natural way. (Right) The same variables (\(47369\)) in the model, with losses computed using our method and \textit{adaptive} sampling of \textit{local structures} (Subsection ~\ref{sec:3.3}), yield the most natural and refined wrinkles and demonstrate superior results compared to uniform sampling when trained for the same number of epochs.}
  \Description{}
  \label{fig:teaser}
\end{teaserfigure}

\input{sec/0_abstract}
\maketitle
  
\input{sec/1_intro}
\input{sec/2_related_work}
\input{sec/3_method}
\input{sec/4_evaluation}
\input{sec/5_limitations_and_conclusion}


\bibliographystyle{ACM-Reference-Format}
\bibliography{sample-base}










\end{document}

%% file: sec/0_abstract.tex
\begin{abstract}
The accurate representation of fine-detailed cloth wrinkles poses significant challenges
in computer graphics. The inherently non-uniform structure of cloth wrinkles mandates the employment of intricate discretization strategies, which are frequently characterized by high computational demands and complex methodologies. Addressing this, the research introduced in this paper elucidates a novel anisotropic cloth regression technique that capitalizes on the potential of implicit neural representations of surfaces. Our first core contribution is an innovative mesh-free sampling approach, crafted to reduce the reliance on traditional mesh structures, thereby offering greater flexibility and accuracy in capturing fine cloth details. Our second contribution is a novel adversarial training scheme, which is designed meticulously to strike a harmonious balance between the sampling and simulation objectives. The adversarial approach ensures that the wrinkles are represented with high fidelity, while also maintaining computational efficiency. Our results showcase through various cloth-object interaction scenarios that our method, given the same memory constraints, consistently surpasses traditional discrete representations, particularly when modelling highly-detailed localized wrinkles.
\end{abstract}

%% file: sec/1_intro.tex
\section{Introduction}
\label{sec:intro}

In recent years, learning-based methods have become increasingly popular for simulating cloth. These methods use neural networks to predict the deformations on virtual garments. A common approach for training these neural networks is supervised learning~\cite{casas2018learning, gundogdu2019garnet, ma2020learning, Bertiche2019, santesteban2019learning}, which requires large amounts of physics-based simulated or animated cloth data as ground truth. The training process minimizes the vertex offsets between the predicted and ground truth meshes. Although inference with these trained networks is nearly real-time, the generalizability of supervised learning methods can be limited and generating sufficient training data can be difficult or time-consuming.

To overcome these limitations, unsupervised learning methods have been developed. Bertiche et al.~\cite{pbns} introduced a novel unsupervised learning method that formulates the loss function as the garment's potential energy. This method jointly trains the neural network weights and evaluates the equations of motion for quasi-static scenarios, allowing the regression of garment vertex positions by directly minimizing the potential energy without the need for training data. Santesteban et al.~\cite{santesteban2022snug} further improved this approach by adding temporal information and kinetic energy to the loss function for dynamic garments, and a hyperelastic material model to characterize in-plane elasticity.

However, these unsupervised techniques demand an explicit representation of the entire garment mesh, leading to extensive networks with slow convergence rates, and low fidelity in representing fine cloth details, e.g., wrinkles. In response, we propose an implicit representation of garments that uses a multi-resolution grid structure. This representation boasts several advantages: reduced memory usage, and most importantly a continuous domain with inherent adaptivity. This adaptivity permits the network weights to capture intricate details at any spatial location without changing the network architecture. Leveraging this strength, we introduce a novel mesh-free sampling technique that reduces reliance on traditional mesh structures. This offers enhanced flexibility and precision in capturing fine cloth details. Employing this sampling approach, we formulated an adversarial loss function, finely-tuned to strike a balance between sampling and simulation objectives, thus aiding in training.

We demonstrate that, under the same memory constraints, our method consistently outperforms traditional discrete representations. This is especially evident in the enhanced simulation results for detailed cloth wrinkles, particularly for small, localized ones.

\paragraph{Contributions.} In summary, the major technical contributions of this paper include
\begin{itemize}
\item A specifically designed multi-resolution grid encoding model for neural implicit surface representation to enable efficient garment simulation.

\item A suitable sampling method specifically designed for adaptive garment simulation.

\item A new formulation for the losses computed on neural implicit surfaces based on a newly proposed sampling local structure.

\item A novel adversarial loss formulation for adaptive garment simulation and its proof of effectiveness.
\end{itemize}

%% file: sec/2_related_work.tex
\section{Related Work}
\label{sec:related_work}
\paragraph{Cloth Simulation.} The simulation of cloth is a long-standing and widely researched topic in computer animation. Since the debut of the seminal Baraff--Witkin model \cite{Baraff1998}, several improvements were proposed to better virtually represent fabrics over the years. These include mixed implicit-explicit solvers \cite{Bridson2005}; improving stability \cite{Choi2005, Thomaszewski2009, Li2015, Kim2020BaraffWitkin}; finite-elements formulations with co-rotational \cite{Etzmuss2003}, hyperelastic \cite{Miguel2016}, linear orthotrophic \cite{Li2015} and Baraff--Witkin \cite{Kim2020BaraffWitkin} energy strains; adaptive remeshing for cloth \cite{Narain2012}, paper \cite{Narain2013} and thin-sheets \cite{Pfaff2014}; efficient modelling of yarn-level fabrics \cite{Cirio2014, Sperl2020, Sperl2021}; anisotropic elastoplasticity coupled with frictional contacts \cite{Jiang2017}, Eulerian-on-Lagrangian contact resolution \cite{Weidner2018}, and sub-millimeter wrinkle synthesis \cite{Wang2021}. For an analysis of different strain formulations along with production implementation practicalities, we refer to Kim and Eberle \cite{kim2020dynamic}.

\paragraph{Wrinkle Simulation.} There has been a significant focus on proficiently enhancing coarse base animations with intricate wrinkle details. Beginning with Grinspun et al. \cite{grinspun2002charms}, who introduced adaptive refinement for wrinkles and folds, the field has progressed with Bergou et al. \cite{bergou2007tracks} utilizing constrained Lagrangian mechanics to mirror low-resolution dynamics. Rohmer et al. \cite{rohmer2010animation} provided dynamic wrinkles integration through strain tensor analysis. Müller and Chentanez \cite{muller2010wrinkle} harnessed position-based dynamics for intricate wrinkles, while Chen et al. \cite{chen2013modeling} emphasized on the interplay of cloth and body, capturing fine wrinkles. Zuenko and Harders \cite{zuenko2019wrinkles}, Rémillard and Kry \cite{remillard2013embedded}, and Casafranca and Otaduy \cite{casafranca2022voronoi} delved into unique methods to replicate human skin wrinkling. Furthermore, tension field theory (TFT) and data-driven approaches, highlighted by works from Chen et al. \cite{chen2021fine} and Wang et al. \cite{wang2010example}, have enriched the field with detailed and realistic wrinkle simulations.

\paragraph{Collision detection.} A crucial step from numerically simulating cloth is the collision detection and response phase. Such process is often the bottleneck of the entire simulation, specially if implemented naively. Since we aim to mimic steps of a physically-based solver during the training phase, it is important to understand how collision detection can be robustly and efficiently implemented on GPUs. Bridson et al. \cite{Bridson2005} adopted the GPU-friendly signed distance functions (SDFs). SDFs were also regressed implicitly by a neural network relative to a given a character pose \cite{Chen2021a}; such an approach can be useful for animated characters, since the majority of the collisions are due to cloth-body interactions. Similarly, Santesteban et al. \cite{Santesteban2021} proposes a self-supervised collision loss that augments decoded network predictions by automatically sampling the latent space connected to a collision loss. Other works also focus on efficiently dealing with cloth self-collisions on the GPU; repulsion-based methods \cite{Stam2009, Macklin2014, fratarcangeli2016vivace, wu2020safe} model spring forces using minimal edge distances to avoid interpenetration. Tang et al. \cite{Tang2018a} implemented an efficient collision-detection algorithm tailored for GPUs that combines spatio--temporal coherence, bounding volume hierarchies, discrete (DCD) and continuous collision detection (CCD). Lastly, Lan et al. \cite{Lan2020} employs a medial axis transform to model volumetric objects, combining spatial hashing and a collision culling algorithm that exploits mathematical properties of the medial axis transform.

\paragraph{Data-driven methods.} Many works have used data-driven methods without relying on Machine Learning, some of which include: example-based wrinkle synthesis \cite{Wang2010}, cloth upsampling for real-time applications \cite{Kavan2011}, efficient mesh representations for clothed humans \cite{Guan2012, Wu2021} and soft tissue animation \cite{Kim2017}. Accurately estimating physical parameters for simulating cloth is an important task in order to faithfully recreate them in virtual environments. Data-driven estimation of cloth parameters include models represented by linear \cite{Wang2011}, Kirchhoff--Love \cite{Miguel2016} and St. Venant--Kirchhoff \cite{Miguel2012, Clyde2017} strain energies.

\paragraph{Machine Learning in Computer Animation.}
Several works \cite{Fulton2019, Tan2019, Chentanez2020, Sanchez-Gonzalez2020, Deng2020, Shen2021} were proposed to reduce computations when regressing physically-based deformations. Tan et al. tailored the computational graph for simulating cloth in both width and depth: a graph-based convolutional neural network encodes the input into a low dimensional space, while a recurrent neural network (RNN) learns a fully differentiable physics loss in a reduced number of iterations. Similarly, but substituting the RNN by a limited set of message passing iterations, deformables \cite{Pfaff2020}, continuous materials \cite{Sanchez-Gonzalez2020}, and soft tissues \cite{Deng2020} were successfully regressed by graph neural networks. The aforementioned approaches, however, only loosely approximate the equations of motion; hence, Fulton et al. \cite{Fulton2019} proposed a subspace solver that directly integrates the the latent space of a non-linear autoencoder to more aggressively reduce the width of the computational graph. Follow up work \cite{Shen2021} identified missing non-linear inertial terms when integrating the latent space of autoencoders. However these terms require third-order (Hessians) network derivatives, which were approximated with a complex-step finite difference method. Other works include modelling cloth--body interactions through point features represented by varying levels of detail \cite{Gundogdu2018}, graph convolutions tailored to cloth regression and upsampling \cite{Chentanez2020}, mapping deformations to a two dimensional spaces to exploit efficient CNN architectures \cite{Jin2018}, high-frequency wrinkle synthesis \cite{Laehner2018a}, decoupling low and high-frequency mesh deformations with mixture models \cite{patel2020tailornet, Zhang2021}. 

%% file: sec/3_method.tex
\section{Method}
\label{sec:method}
We propose a novel representation of the garment surface using implicit neural representations; the details of the surface are captured using neural network parameters. Building on this implicit neural representation, we introduce a new formulation to compute the simulation losses based on a \textit{sampling local structure}. We propose a minimax adversarial objective function. During training, we alternate between sampling and simulation objectives to strike a balance between speed and accuracy.

\paragraph{Structure.} In Subsection \ref{sec:3.1} we detail our approach to utilizing neural networks for representing the implicit surface, which includes our specially designed multi-resolution grid encoding neural network model. In Subsection \ref{sec:3.2}, we delve into the sampling method and explain the rationale behind our choice. Subsection \ref{sec:3.3} introduces a novel loss computation method for the neural implicit surface, based on \textit{sampling local structures}. Subsection \ref{sec:3.4} presents our innovative minimax adversarial loss formulation, complete with algorithm details.

\subsection{Representation of Surfaces}
\label{sec:3.1}
The traditional representations like the mass-spring system or the finite element method necessitate the discretization of the garment surface. Capturing intricate details, such as cloth wrinkles, with these discretized surfaces is often challenging unless extremely high resolutions are used, which in turn increases computational costs. As an alternative, we employ an implicit neural representation for the cloth. This method provides a continuous domain with inherent adaptivity. Our study emphasizes quasi-static scenarios, as our main objective is to represent cloth behavior accurately and stably in situations with minimal dynamic changes.

To parameterize the shape of a cloth, we use the UV coordinates. This is formally represented by the function $\mathcal{S}$:
\begin{equation}
\begin{aligned}
\mathcal{S}: \mathbb{R}^2 &\rightarrow \mathbb{R}^3, \\
\mathbf{p}_{UV} &\mapsto \mathbf{p}_{3D},
\end{aligned}
\end{equation}
where $\mathbf{p}_{UV}$ represent the UV coordinates, and $\mathbf{p}_{3D}$ represent the deformed 3D position.

Each 3D position on the deformed cloth shape can be decomposed into two components, the undeformed position $\mathbf{p}_0$, and the 3D deformation $\Delta \mathbf{p}_{3D}$ on top of the undeformed position, i.e., 
\begin{equation}\mathbf{p}_{3D} = \mathbf{p}_0 + \Delta \mathbf{p}_{3D}.\end{equation}

Given that the undeformed position $\mathbf{p}_0$ is known, our primary objective becomes learning the deformation $\Delta \mathbf{p}_{3D}$. This deformation is captured by the function $\mathcal{D}$:
\begin{equation}
\begin{aligned}
\mathcal{D}: \mathbb{R}^2 &\rightarrow \mathbb{R}^3, \\
\mathbf{p}_{UV} &\mapsto \Delta \mathbf{p}_{3D},
\end{aligned}
\end{equation}
where $\Delta \mathbf{p}_{3D}$ is the difference in the 3D position due to deformation.

Following the reasoning above, in our implicit neural representation, instead of using a neural network to represent the map $\mathcal{S}$, we opt to using a neural network to represent the map $\mathcal{D}$. The strategy of incremental learning — as exemplified by ResNets \cite{he2016deep} in learning residuals — offers distinct advantages, particularly when applied to the task of modeling 3D shapes. When a network is focused on capturing the nuanced differences from a base structure, it inherently grapples with simpler and often smaller magnitudes of change compared to recreating an intricate shape in its entirety. This eases the learning process, making the optimization landscape less fraught with local minima that could trap the model in sub-optimal solutions. Furthermore, this incremental approach can act as an implicit form of regularization. Instead of the expansive freedom to generate any conceivable shape, which could inadvertently lead to overfitting, the model is gently tethered to a foundational shape, adapting and molding it through subtle deformations.

For training the network, we set the physically based energies as the losses, and utilize back-propagation to optimize the network parameters, this can let us directly obtain the 3D deformation of the garments without explicitly computing the forces in the physical system for solving the equation of motion.

\paragraph{Multi-resolution Grid Encoding Model.} In computer graphics, the concept of the UV domain refers to a two-dimensional coordinate system that is integral to texture mapping on 3D surfaces. Each vertex of a 3D model is linked with a 2D coordinate $(u, v)$ that determines its correspondence on the texture. This UV mapping effectively transforms a 3D surface into a two-dimensional representation. Because of this, the UV parameterized domain inherently possesses spatial properties. Points that are adjacent or near each other in UV space often have a similar proximity on the actual 3D model.

This spatial characteristic of the UV domain is not just theoretical; it provides actionable insights. By understanding how the UV space spatially correlates with the 3D model, this knowledge can be integrated into the encoding process. Such integration of prior knowledge can significantly enhance the efficiency and accuracy of encoding, tailoring it more closely to the nuances of the 3D model it represents.

At the heart of this enhanced encoding is the concept of multi-resolution grid encoding. Think of this as viewing a picture with varying levels of zoom. At a lower resolution or a more zoomed-out view, you see broader features, capturing the overall essence. Conversely, a high-resolution or zoomed-in perspective reveals the minute intricacies. This method is pivotal for systems where spatial relationships exist in a hierarchical manner. The vast world of garment simulation provides an apt illustration. Here, while the broad shape of a shirt or a dress is an overarching spatial feature, the fine stitches, textures, or minute wrinkles are the granular details. The multi-resolution approach ensures both these details are captured and represented with fidelity.



In our model that is specifically designed for this garment simulation, in order to produce a more standardized representation, we employ bilinear interpolation as a means of embedding unstructured texture coordinates into a structured grid. This procedure encodes local topological information into the neural network. In details, each UV point $\textbf{p}_{UV} = (x, y)$ is passed into the GE (Grid Encoding) to obtain the bilinearly interpolated grid features on each layer of the multi-resolution grid. Such multi-resolution grid is constructed of $L$ layers, where $L$ is a user-defined constant. Suppose the densest layer is of resolution $N_{\text{max}}$, then the rest layers are of resolution 
$\lfloor N_{\text{max}} / 2^1 \rfloor, \lfloor N_{\text{max}} / 2^2 \rfloor, \cdots, \lfloor N_{\text{max}} / 2^L \rfloor$.
Note that here we assume 
$L \leq \lfloor \log_2 N_{\text{max}} \rfloor$.

In details, on layer \( l \), the interpolated feature vector \( \boldsymbol{\mathcal{F}}^l(x, y) \) can be computed as:
\begin{equation}
\begin{aligned}
\alpha &= \frac{1}{(x_2 - x_1)(y_2 - y_1)}, \\
\boldsymbol{v_x} &= \begin{bmatrix} x_2 - x & x - x_1\end{bmatrix}, \\
\boldsymbol{M} &= \begin{bmatrix} \boldsymbol{\mathcal{F}}^l(x_1, y_1) & \boldsymbol{\mathcal{F}}^l(x_1, y_2) \\ \boldsymbol{\mathcal{F}}^l(x_2, y_1) & \boldsymbol{\mathcal{F}}^l(x_2, y_2)\end{bmatrix}, \\
\boldsymbol{v_y} &= \begin{bmatrix} y_2 - y \\ y - y_1 \end{bmatrix}, \\
\boldsymbol{\mathcal{F}}^l(x, y) &= \alpha \cdot \boldsymbol{v_x} \cdot \boldsymbol{M} \cdot \boldsymbol{v_y}
\end{aligned}
\end{equation}
where \( x_1 = \lfloor x \rfloor \), \( x_2 = x_1 + 1 \), \( y_1 = \lfloor y \rfloor \), and \( y_2 = y_1 + 1 \).

And then these grid features on each layer are concatenated together to form the input to the MLP,
\begin{equation}
\mathrm{\mathbf{GE}} (x, y) = \boldsymbol{\mathcal{F}} (x, y) = \boldsymbol{\mathcal{F}}^1 (x, y) \oplus \boldsymbol{\mathcal{F}}^2 (x, y) \oplus \cdots \oplus \boldsymbol{\mathcal{F}}^L (x, y),
\end{equation}
where $L$ is the total number of layers. Then, we pass this $\mathcal{F}(x, y)$ through the MLP, and the output of the MLP represents the 3D deformation:
\begin{equation}
\Delta \mathbf{p}_{3D} = \mathrm{\mathbf{MLP}} \left(\mathrm{\mathbf{GE}} (x, y) \right).
\end{equation}

\begin{figure*}[!htb]
  \centering
  \includegraphics[width=1.0\linewidth]{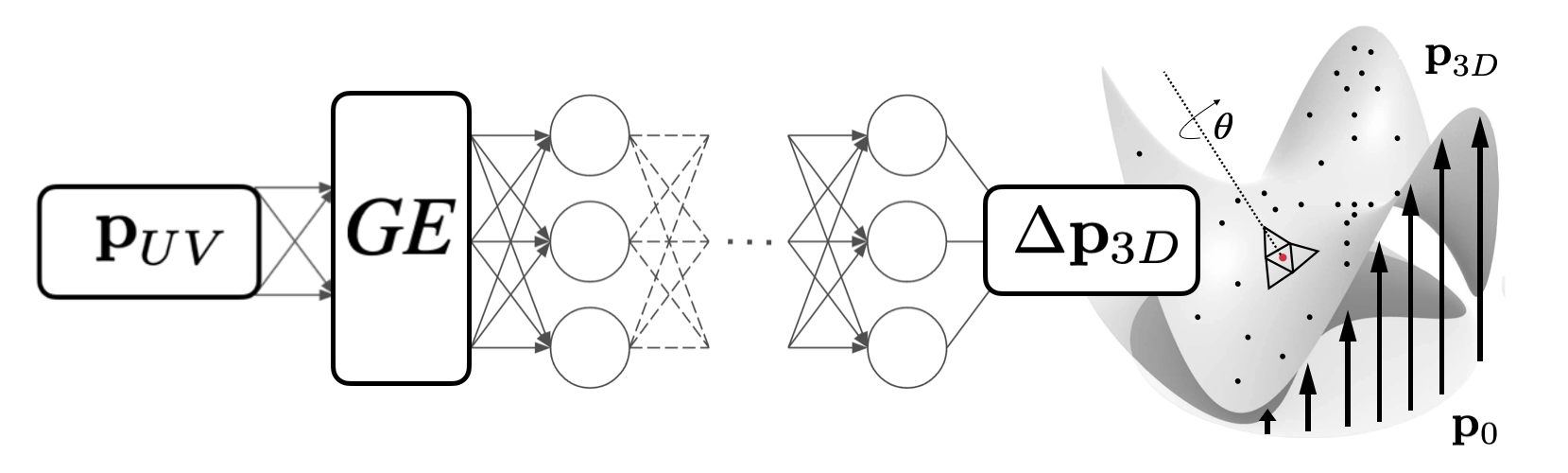}
  \caption{
  \label{fig:GE}
\textbf{The pipeline.} The input to the pipeline consists of two-dimensional (2D) UV coordinates $\mathbf{p}_{UV}$. It is first encoded using multi-resolution grid encoding (GE) to obtain the bilinearly interpolated grid features. These features are then fed into the MLP, and the output of the MLP is $\mathbf{p}_{3D}$, representing the three-dimensional (3D) deformation. Suppose a point in the undeformed state is named $\mathbf{p}_0$; then the deformed 3D position of such a point is $\mathbf{p}_{3D} = \mathbf{p}_0 + \Delta \mathbf{p}_{3D}$. The losses are computed using our novel method on the \textit{sampling local structures} which were constructed in 2D and then mapped to 3D atop the learned 3D neural implicit surface.}

\end{figure*}

We illustrate the pipeline containing the multi-resolution grid encoding model in Figure ~\ref{fig:GE}. In our implementation, the number of layers in the multi-resolution grid, the number of features on each grid cell, the resolution of the grid cells, and the number of layers and sizes of the following MLP are all user-definable. We will provide a detailed description of the architecture used in our experiments in Section ~\ref{sec:evaluation}.

\subsection{Sampling Method}
\label{sec:3.2}
To leverage the advantages of the continuous domain and the adaptive benefits of implicit neural representations, we investigate sampling methods specifically designed for the parameterized UV space. This ensures a denser concentration of sampling points in regions with intricate details. In this specific case, we assume that the UV parametrization for the garment has minimal distortion. Since garments are often designed using developable surfaces, it can be easily cut into pieces and laid flat on a plane.

In every optimization step, we select points based on a probability distribution. Regions with more intricate details have higher probability values. However, understanding the genuine continuous probability density function (PDF) can be challenging. Still, there are strategies to address this. One of the strategies is to create a discrete approximation of the elusive PDF and select sampling points based on this approximation.

\paragraph{Probability Computation.} To better grasp and represent a continuous, unseen PDF, we approximate its values at select discrete points. This snapshot forms a discrete model of the actual continuous PDF, allowing for clearer visualization and simplified sampling. Assuming the true PDF is continuous and smooth, this discrete version is often high fidelity. This is because in a smooth function, closely situated points have similar values. Thus, the values we determine at these discrete locations are likely reliable indicators of the continuous function's behavior in their immediate vicinity. We divide the 2D UV space into a grid of moderate density. For each grid cell indexed as $(i, j)$, we calculate a weighted sum of the losses at the center point within that cell. We assume that the probability value $p_{ij}$ of the grid at this specific epoch is represented by this weighted sum $\hat{p}_{ij}$. Starting from a uniform discrete PDF in the first epoch, we update the sampling PDF in each subsequent epoch to align it more closely with the estimated PDF for that specific epoch using linear interpolation:
\begin{equation}
p'_{ij} = \gamma p_{ij} + (1 - \gamma) \hat{p}_{ij},
\end{equation}
where $p'_{ij}$ is the probability value in the next epoch, and $\alpha$ is a user-defined constant. Next, we determine an appropriate scaling to produce a discrete PDF so that all the values sum up to $1$. Details on computing the losses are provided in Subsection \ref{sec:3.3}.


\paragraph{Inverse Transform Sampling.} When dealing with a 2D discrete probability density function (PDF), inverse transform sampling becomes a crucial tool for sampling points. Imagining a 2D discrete space where each point is defined by coordinates \((i, j)\), every point is assigned a specific probability, which we will represent as \(p_{ij}\).

The first stage in the inverse transform sampling process is the calculation of the marginal PDF for each row. This is achieved by taking the sum of probabilities along each row. If you imagine an array or matrix, it is akin to summing up all values in a specific row. We can express the marginal PDF of a given row \(i\) as \(p_i\), represented mathematically by the formula:
   \begin{equation}p_i = \sum_{j=1}^{N} p_{ij}, \text{ for } i \in [M],\end{equation}
where \(M\) stands for the total rows and \(N\) symbolizes the total columns.

Once the marginal PDF is determined, the next phase is deducing the marginal cumulative density function (CDF) for each row. This involves cumulatively summing the probabilities of rows up to a given point. For any given row \(i\), the marginal CDF is notated as \(P_i\), and it's calculated as:
\begin{equation}P_i = \sum_{k=1}^{i} p_{k}, \text{ for } i \in [M].\end{equation}

With the marginal CDF in place, the next move is to generate a random number \(u\), sourced from a uniform distribution in the range \([0, 1]\). This number plays a pivotal role as it will guide us in identifying the sampled row index. Essentially, we are looking for the smallest row index \(i\) where \(p_i\) either equals or surpasses \(u\), mathematically put as:
\begin{equation}i = \inf \{k : P_k \geq u\}.\end{equation}

Having pinpointed the row, we then dive deeper into it and compute its conditional CDF. This requires summing up the conditional probabilities along that specific row. For the chosen row \(i\) and any column \(j\), the column-wise CDF is represented as \(Q_{ij}\) and is computed via:
\begin{equation}Q_{ij} = \sum_{l=1}^{j} \frac{p_{il}}{p_i}, \text{ for } j \in [N].\end{equation}

The last steps of the process are quite similar to the earlier ones but on a columnar basis. A random number \(v\) is pulled from a uniform distribution within the range \([0, 1]\), directing us to the specific column index to be sampled within our earlier chosen row. We determine \(j\) by pinpointing the smallest column index such that \(Q_{ij}\) equals or surpasses \(v\), represented as:
\begin{equation}
j = \inf \{l : Q_{il} \geq v\}.
\end{equation}

By the end of this process, we obtain a randomly sampled point \((i, j)\). We
then randomly sample a point within the grid corresponding to this pair of indices. This point aligns with the original two-dimensional distribution mapped out by \(p_{ij}\). An essential thing to remember is that for the entire process to be accurate and valid, the probabilities \(p_{ij}\) must be normalized, ensuring their sum equals $1$.

\paragraph{Lloyd's Relaxation.} Direct sampling according to the PDF may result in points that are overly concentrated in specific regions. To address this, we use Lloyd's Relaxation on points acquired through inverse transform sampling. Lloyd's Relaxation is a critical process in ensuring a balanced and uniform distribution of points within a defined space, especially when direct sampling in line with the PDF might lead to an undesired concentration of points in certain regions. This method is primarily employed to refine the positions of points acquired through inverse transform sampling.

The principle behind this technique is the optimization of point positions to improve their distribution in relation to the Voronoi diagram. Imagine we have an initial set of points, which we can denote as \(\mathcal{P} = \{\mathbf{p}_1, \mathbf{p}_2, \ldots, \mathbf{p}_n\}\). Each point, say \(\mathbf{p}_i\), has coordinates represented as \( (x_i, y_i)\) corresponding to the \(i\)-th point.

To better understand how Lloyd's Relaxation functions, we walk through the steps in a 2D setting. The process commences by constructing the Voronoi diagram using the present positions of the points in the set \(\mathcal{P}\). This is a spatial division of a plane where each division (or region) contains points that are closest to a specific point in set \(\mathcal{P}\).

Upon the construction of the Voronoi diagram, the next step involves calculating the centroid for each point \(\mathbf{p}_i\) within the set \(\mathcal{P}\). The centroid, \(\mathbf{c}_i\), represents the average coordinates of all points lying inside the Voronoi region corresponding to \(\mathbf{p}_i\). Mathematically, the centroid can be expressed as:
\begin{equation}\mathbf{c}_i = \frac{1}{m_i}\sum_{\mathbf{q}_j \in R_i}\mathbf{q}_j,\end{equation}
where \(R_i\) symbolizes the Voronoi region related to \(\mathbf{p}_i\), and \(m_i\) denotes the count of points within that specific region.

Following the centroid calculations, each point \(\mathbf{p}_i\) has its position updated to match the coordinates of its respective centroid, \(\mathbf{c}_i\).

This entire sequence of steps is repeated either for a pre-defined number of iterations or until certain convergence criteria are achieved. The beauty of Lloyd's Relaxation is that as these steps are performed iteratively, the points in set \(P\) progressively shift toward a configuration that is more evenly spaced, thereby optimizing the Voronoi diagram. This results in a more uniform distribution of points, avoiding the problem of concentration in specific regions.

\subsection{Simulation Losses}
\label{sec:3.3}
To harness the distinct advantages of the implicit neural representation and potentially delve into its adaptivity, we redefined the simulation energies tailored for this implicit neural representation. We achieved this by constructing local sampling structures atop our neural implicit surface. Using these sampling local structures, we can compute the losses for the corresponding sampling point, based on the relative positions of the vertices within the local structure.

\begin{figure}[!htb]
  \centering
  \includegraphics[width=1.0\linewidth]{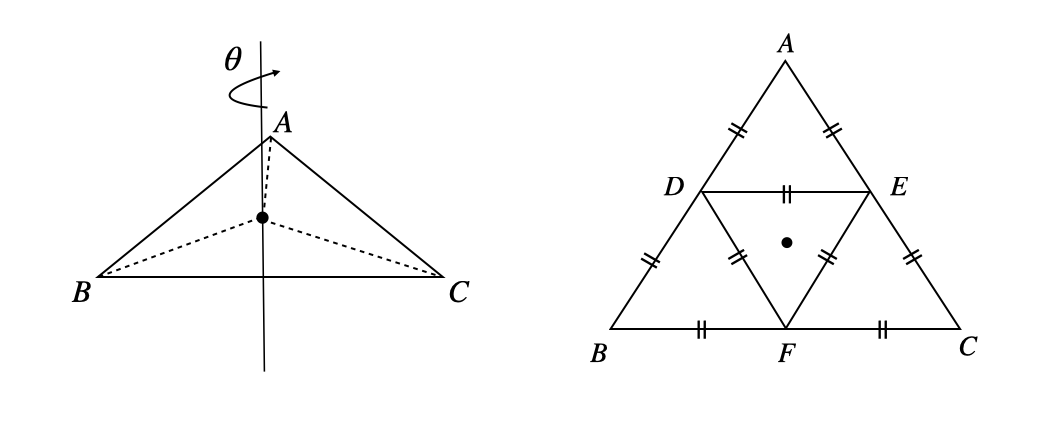}
  \caption{
  \label{fig:structure}
  Sampling Local Structure}
\end{figure}

For clarity, we use the term \textit{3D sampling points} to refer to the 3D points corresponding to the sampling points in the UV space. The original sampling points in the UV space are referred to as \textit{2D sampling points}. For each 2D sampling point, we construct four equilateral triangles around it in the UV space, as shown in Figure \ref{fig:structure}. The triangle $ABC$ has a degree of freedom $\theta$ which denotes the in-plane rotation. This $\theta$ is a randomly generated number within the range $[0, 2\pi / 3]$ in each epoch.

It is important to note that all 2D sampling points within the sampling local structures are initially established in the UV space, then they are mapped from 2D to 3D. Remember, the 3D position, \(\mathbf{p}_{3D}\), corresponding to a 2D UV point, \(\mathbf{p}_{UV} = (x, y)\), can be easily computed using the deformation network:
\begin{equation}
\mathbf{p}_{3D} = \mathbf{p}_{0} + \mathrm{\mathbf{MLP}} \left(\mathrm{\mathbf{GE}} (x, y) \right),
\end{equation}
where \(\mathbf{p}_{0}\) signifies the 3D position corresponding to the UV point in its undeformed state. We further analyze the local surface properties based on this 3D sampling local structure.

The local sampling structure presents several noteworthy benefits:

Firstly, in physics-based simulations that utilize traditional mesh representation, the outcome of the simulation can be heavily influenced by the quality of the triangulation. However, preparing the input model as a mesh with good triangulation quality often requires meticulous attention and intricate mesh processing algorithms. Each of our local sampling structure is locally Delaunay in 2D space by construction. Moreover, garments are often designed using developable surfaces which can be readily segmented and flattened on a plane. As a result, when mapping the local sampling structure to 3D space, only minimal distortion occurs. Based on these premises, the 3D local sampling structure usually retains a high-quality local triangulation.

Secondly, the technique of sampling local structures with random orientations offers a nuanced way to comprehend garment material behavior. Instead of relying solely on traditional mesh-based representations, this approach focuses on the minuscule, localized structures within the material. In doing so, it is not limited to a single orientation or direction. By randomly sampling these structures, the method accounts for losses in strain and bend from various angles. This is invaluable for understanding garments, as it sheds light on how the material reacts when worn, especially during movement. Many such materials are anisotropic, exhibiting properties that vary depending on the direction. For instance, some fabrics may stretch more in one direction than another. This contrasts with isotropic materials, which display consistent properties irrespective of direction. Given these differences, the sampling method is particularly suitable for simulating anisotropic garments. Instead of assuming uniformity, it samples various orientations of local structures, capturing the unique attributes of anisotropic materials.

In the remainder of this subsection, we will demonstrate how we define the losses using this innovative sampling of the local structure.
\subsubsection{Strain Loss} The computation of strain loss consists of three parts: precomputation, rest length computation, and the loss computation itself.
\paragraph{Precomputation.} Let us consider a garment mesh, denoted as $\mathcal{M}$, along with its corresponding UV parametrization $\phi: \mathbb{R}^3 \rightarrow [0, 1]^2$. To represent the 3D positions of the mesh vertices, we employ a square image in the range of $[0, 1]^2$. Specifically, we encode the \textit{scaled} 3D vertex positions as RGB values and assign them to the corresponding pixels of the image. Alternatively, an RGBA image can be used, where the additional \textit{Alpha} channel can represent a mask. Users have the flexibility to specify the resolution of the image, with a default value of 1024.

Let $\mathbf{p}_1$, $\mathbf{p}_2$, and $\mathbf{p}_3$ denote the 3D positions of three vertices within the mesh $\mathcal{M}$. It is possible to determine the 3D position of any 2D point in the UV space, provided that it lies within the convex hull defined by $\phi(\mathbf{p}_1)$, $\phi(\mathbf{p}_2)$, and $\phi(\mathbf{p}_3)$. This interpolation is achieved using the Barycentric interpolation method. It is noteworthy that this step only needs to be computed once for each garment mesh in its rest pose, and parallel computation techniques can be employed to minimize the computational time required.

\paragraph{Barycentric Interporlation.}


Consider a 2D triangle defined by vertices $\mathbf{p}_1$, $\mathbf{p}_2$, and $\mathbf{p}_3$. Any point $\mathbf{p}$ within this triangle can be represented as a unique linear combination of these vertices:
\begin{equation}
\mathbf{p} = \lambda_1\mathbf{p}_1 + \lambda_2\mathbf{p}_2 + \lambda_3\mathbf{p}_3,
\end{equation}
where $\lambda_1$, $\lambda_2$, and $\lambda_3$ are the Barycentric coordinates of $\mathbf{p}$, and $\lambda_1 + \lambda_2 + \lambda_3 = 1$.

These coordinates do not only depict the weights of the vertices for interpolating $\mathbf{p}$, but they also remain invariant under affine and barycentric transformations. This invariance yields consistent interpolations under transformations, providing a unique advantage over other interpolation methods.

Using Barycentric coordinates, we can express Barycentric interpolation in the form of:
\begin{equation}
F(\mathbf{p}) = \lambda_1F(\mathbf{p}_1) + \lambda_2F(\mathbf{p}_2) + \lambda_3F(\mathbf{p}_3),
\end{equation}
where $F$ represents the function that we wish to interpolate (such as color, texture, or other attributes), and $F(\mathbf{v}_i)$ denotes the attribute value at vertex $\mathbf{v}_i$. Note that in our case, $F$ is the \textit{inverse} of UV parametrization function $\phi$, under the assumption that $\phi$ is bijective and thus invertible.

\begin{figure}[!htb]
  \centering
   \mbox{} \hfill
  \includegraphics[width=0.5\linewidth]{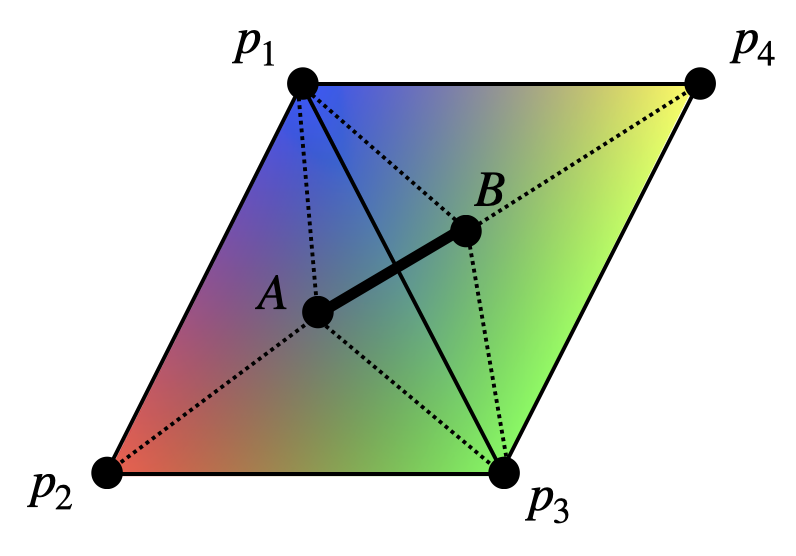}
   \hfill \mbox{}
  \caption{
  \label{fig:rest_length}
  Rest length computation for hypothetical edge $AB$}
\end{figure}

\paragraph{Rest Length Computation.} We use the term \textit{valid} to refer to 2D points that lie within the union set of the 2D UV triangulation. By utilizing Barycentric interpolation on the pre-computed 3D positions in the rest pose, it becomes possible to compute the 3D position of any \textit{valid} point within the 2D UV space. For instance, considering two such 2D points denoted as $\mathbf{A}$ and $\mathbf{B}$, as depicted in Figure \ref{fig:rest_length}, an approximation of the rest length of the hypothetical edge connecting the two 3D points represented by $\mathbf{A}$ and $\mathbf{B}$ can be determined using the Euclidean norm, expressed as 
\begin{equation}\|\phi^{-1}(\mathbf{A}) - \phi^{-1}(\mathbf{B})\|_2.\end{equation} 
It is necessary to compute the 3D edge length between any two 3D points represented by two \textit{valid} points since the vertices of the sampling triangles can be any \textit{valid} 2D points.

\paragraph{Loss Computation.} The strain loss is the potential elastic energy of the system, formulated based on the Hooke's law in the mass spring system to ensure that the cloth is not excessively stretched or compressed. 
\begin{figure}[!htb]
  \centering
   \mbox{} \hfill
  \includegraphics[width=0.7\linewidth]{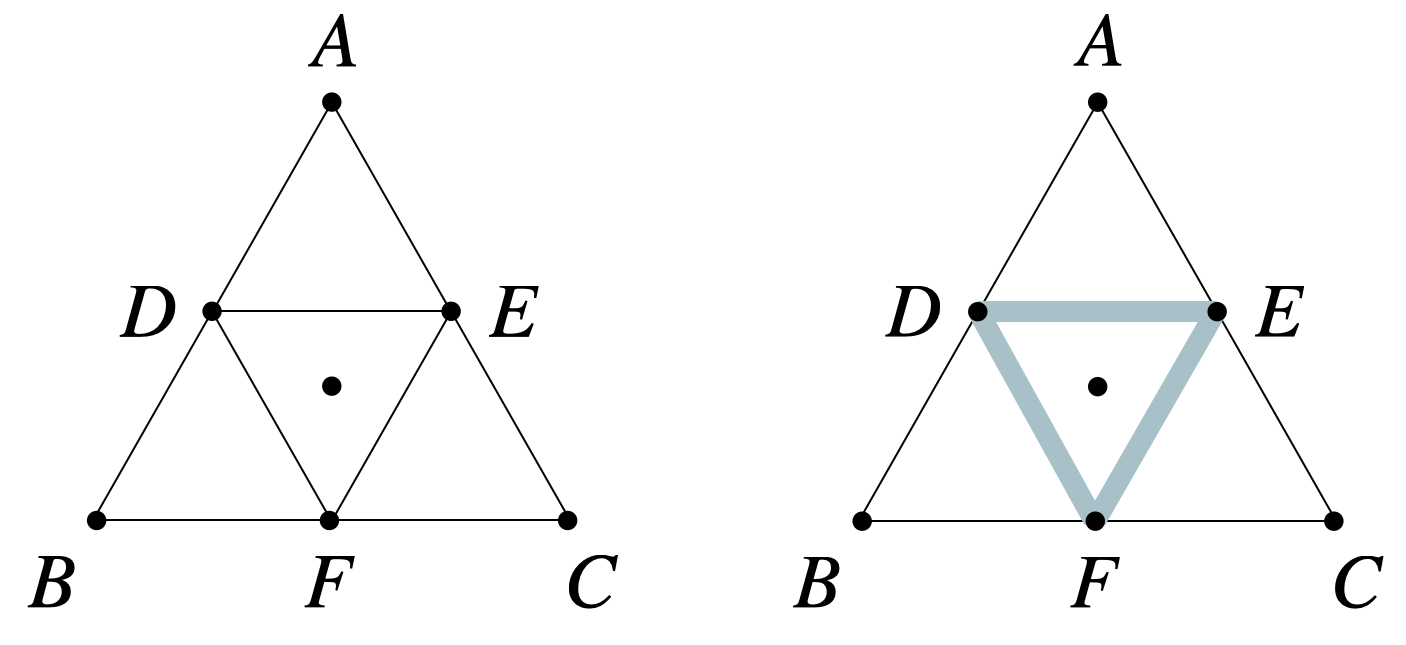}
   \hfill \mbox{}
  \caption{
  \label{fig:strain}
  Strain Loss: Immediate Edge Set}
\end{figure}
Let $E(\mathbf{p}, \boldsymbol{\Theta})$ be the immediate edge set of sampling point $\mathbf{p}$ when the surface is in the state captured by parameter $\boldsymbol{\Theta}$, with its edges marked in blue in Figure \ref{fig:strain}, for a given sampling local structure. Then, the strain loss at the 3D sampling point $\mathbf{p}$ can be computed in two ways. The first way is formulated as:
    \begin{equation}
    \mathcal{L}_\text{Strain} (\textbf{p}, \boldsymbol{\Theta}) = \sum \limits_{\mathbf{e} \in E(\textbf{p}, \boldsymbol{\Theta})} (\|\mathbf{e}'\|_2 - \|\mathbf{e}\|_2)^2,
    \end{equation}
    which weights more heavily for edges with larger edge lengths. However, in the second formulation, the strain energy is computed according to the ratio of the length change and the original edge length:
    \begin{equation}
    \mathcal{L}_\text{Strain}(\textbf{p}, \boldsymbol{\Theta}) = \sum \limits_{\mathbf{e} \in E(\textbf{p}, \boldsymbol{\Theta})} \left(\frac{\|\mathbf{e}'\|_2 - \|\mathbf{e}\|_2}{\|\mathbf{e}\|_2}\right)^2,
    \end{equation}
    which weights equally for edges with different edge lengths. We opt for the second formulation in our implementation.

    The total strain loss for all the 3D sampling points is then computed as:
    \begin{equation}
    \mathcal{L}_\text{Strain} (\mathcal{P}, \boldsymbol{\Theta}) = \sum \limits_{\textbf{p} 
    \in \mathcal{P}}\sum \limits_{\mathbf{e} \in E(\textbf{p}, \boldsymbol{\Theta})} \left(\frac{\|\mathbf{e}'\|_2 - \|\mathbf{e}\|_2}{\|\mathbf{e}\|_2}\right)^2,
    \end{equation}
    where $\mathcal{P}$ represents the set of all 3D sampling points.
    
\subsubsection{Bend Loss}
\begin{figure}[!htb]
  \centering
   \mbox{} \hfill
  \includegraphics[width=0.9\linewidth]{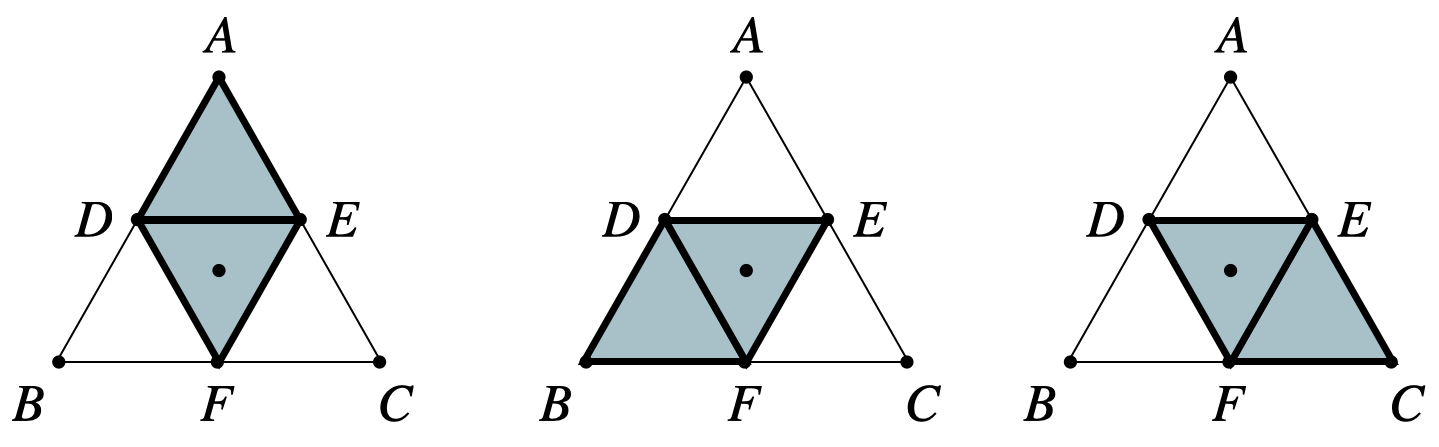}
   \hfill \mbox{}
  \caption{
  \label{fig:bending}
  Bend Loss: Adjacent Face Pairs}
\end{figure}

The bending loss penalizes differences between neighbouring face normals, effectively enforcing locally smooth surfaces. Given a sampling local structure constructed around the sampling point $\mathbf{p}$ on a neural implicit surface captured by parameter $\boldsymbol{\Theta}$, let the set of the face pairs be $\mathcal{FP}(\textbf{p}, \boldsymbol{\Theta})$, for each face pair $\{f_1, f_2\} \in \mathcal{FP} (\textbf{p}, \boldsymbol{\Theta})$, we denote the corresponding \textit{normalized} face normals as $\{\mathbf{n}_1, \mathbf{n}_2\}$. There are three such face pairs in each sampling local structure, as shown in Figure \ref{fig:bending}. Let $k_b$ be the bending constant, $\bar{A}$ be the area sum of the two incident faces, and let $\mathbf{e}_0$ represent the edge connecting the two faces. Then the bending loss at the 3D sampling point $\mathbf{p}$ can be formulated as
    \begin{equation}
    \mathcal{L}_\text{Bend}(\textbf{p}, \boldsymbol{\Theta}) = \sum \limits_{\{f_1, f_2\} \in \mathcal{FP}(\textbf{p}, \boldsymbol{\Theta})} \frac{1}{2} \cdot k_b \cdot \frac{\sqrt{3} \|\mathbf{e}_0\|_2^2}{2 \bar{A}} \cdot \|\mathbf{n}_1 - \mathbf{n}_2\|_2^2.
    \end{equation}

In our case, we want to ignore the scale difference in sampling local structures to ensure that each sampling point is weighted the same. The bend loss computation can instead be formulated as:
    \begin{equation}
    \mathcal{L}_\text{Bend}(\textbf{p}, \boldsymbol{\Theta}) = \sum \limits_{\{f_1, f_2\} \in \mathcal{FP(\textbf{p}, \boldsymbol{\Theta})}} \|\mathbf{n}_1 - \mathbf{n}_2\|_2^2,
    \end{equation}

where all the constants are absorbed into the weight for the bend loss in the weighted sum.

The total bend loss for all the 3D sampling points can thus be calculated as
\begin{equation}
    \mathcal{L}_\text{Bend} (\mathcal{P}, \boldsymbol{\Theta})= \sum \limits_{\textbf{p} 
    \in \mathcal{P}} \sum \limits_{\{f_1, f_2\} \in \mathcal{FP}(\textbf{p}, \boldsymbol{\Theta})} \|\mathbf{n}_1 - \mathbf{n}_2\|_2^2,
\end{equation}
where $\mathcal{P}$ represents the set of all 3D sampling points.

\subsubsection{Gravity Loss}
We also incorporate a term that aims to generate more realistic garment predictions by modeling the effect of gravity. Based on classical mechanics, the potential gravitational energy of a 3D sampling point $\mathbf{p}$ in a surface state captured by parameter $\boldsymbol{\Theta}$ can be calculated as
    \begin{equation}
    \mathcal{L}_\text{Gravity}(\textbf{p}, \boldsymbol{\Theta}) = m(\textbf{p}) \cdot g \cdot h(\textbf{p}, \boldsymbol{\Theta}),
    \end{equation}
    where $m(\mathbf{p})$ is the mass of a 3D sampling point $\mathbf{p}$, $g$ is the gravitational acceleration, and $h(\textbf{p}, \boldsymbol{\Theta})$ is the height of $\mathbf{p}$ measured in a user specified axis, note that by default the axis of gravity is set to be the $z$ axis in our implementation.

The total gravity loss is computed by summing up over all the 3D sampling points:
    \begin{equation}
    \mathcal{L}_\text{Gravity}(\mathcal{P}, \boldsymbol{\Theta}) = \sum_{\mathbf{p} \in \mathcal{P}}m(\textbf{p}) \cdot g \cdot h(\textbf{p}, \boldsymbol{\Theta}),
    \end{equation}
where $\mathcal{P}$ represent the set of all 3D sampling points.

\subsubsection{Collision Loss}
The model needs to handle collisions with other objects. To do so, we design the following loss:
    \begin{equation}
    \mathcal{L}_{\text {Collision}} (\textbf{p}, \boldsymbol{\Theta}) = \sum_{(i, j) \in \mathcal{A}(\textbf{p}, \boldsymbol{\Theta})} \min \left(\mathbf{d}_{j, i} \cdot \mathbf{n}_j-\epsilon, 0\right)^2,
    \end{equation}
    where $\mathcal{A}(\textbf{p}, \boldsymbol{\Theta})$ represents the set of correspondences $(i, j)$ between all the 3D vertices in the local structure sampled around the point $\textbf{p}$ on a surface parameterized by $\boldsymbol{\Theta}$, and the colliding object, respectively. These correspondences are found using nearest neighbors. $\mathbf{d}_{j, i}$ is the vector that goes from the $j$-th vertex of the colliding object to the $i$-th vertex of the outfit, $\mathbf{n}_j$ represents the normal vector at the $j$-th vertex of the colliding object, and $\epsilon$ is a small positive threshold used to enhance robustness.
The total collision loss is computed by summing up over all the 3D sampling points in $\mathcal{P}$:
        \begin{equation}
    \mathcal{L}_{\text {Collision}} (\mathcal{P}, \boldsymbol{\Theta}) = \sum_{\textbf{p} \in \mathcal{P}} \sum_{(i, j) \in \mathcal{A}(\textbf{p}, \boldsymbol{\Theta})} \min \left(\mathbf{d}_{j, i} \cdot \mathbf{n}_j-\epsilon, 0\right)^2.
    \end{equation}
    This loss term is vital for ensuring that the predictions of the garment are valid, as its gradients will encourage the vertices of the outfit to move away from the colliding object.

\subsection{Adaptivity}
\label{sec:3.4}
\subsubsection{Minimax Adversarial Loss Formulation}
Drawing from the methods discussed previously, we can now outline our approach to constructing the adaptive sampling framework. During each epoch, we prioritize sampling points in areas with finer details and then adjust the network's weights to update the neural implicit surface. Ideally, if we could sample an infinite number of points in each epoch, we would compute the losses at each of these points and subsequently update the network weights using them all.

Let us denote the sampling point as $\mathbf{p} = (x, y)$ and the parameters of the implicit neural surface as $\boldsymbol{\Theta}$. Suppose the sampling space is $[0, 1]^2$. Let $\mathcal{F}= \{\text{Strain, Bend, Gravity, Collision}\}$ represent the set of loss names. We can formulate the ideal optimization problem as:
\begin{equation}
\begin{aligned}
\min_{\boldsymbol{\Theta}} \int_0^1 \int_0^1 \sum_{f \in \mathcal{F}} \mathcal{W}_f \mathcal{L}_{f} ((x, y), \boldsymbol{\Theta}) \,\,\, dx \, dy,
\end{aligned}
\end{equation}
where $\mathcal{W}_f$ represents the corresponding loss weight for the loss named $f \in \mathcal{F}$ in the weighted sum.

However, due to memory and time constraints, infinite sampling is not feasible. We are limited to a finite number of sample points within the domain for each iteration.

The question then arises: \textit{where should these points be sampled?} A straightforward strategy is to uniformly sample within the sampling space. However, this approach may not be efficient enough, as uniform sampling may not prioritize sampling regions that require more attention. Therefore, based on this consideration, we propose a strategy to sample more densely in regions with finer details. This corresponds to areas where the losses are higher, and it will be more efficient than a simple uniform sampling method in cases where small regions need more attention than others, such as cloth wrinkles. Based on this strategy, assuming the set of the sampling points is $\mathcal{P} = \{\mathbf{p}_1, \mathbf{p}_2, \cdots, \mathbf{p}_N\}$, $N \in \mathbb{N}$. Let $\mathcal{F}= \{\text{Strain, Bend, Gravity, Collision}\}$ represent the set of loss names, we formulate our heuristic optimization problem as:
\begin{equation}
\min_{\boldsymbol{\Theta}} \max_{\mathcal{P}} \sum_{\mathbf{p} \in \mathcal{P}} \sum_{f \in \mathcal{F}} \mathcal{W}_f \mathcal{L}_f(\mathbf{p}, \boldsymbol{\Theta}),
\end{equation}
where $\mathcal{P}$ follows some constraints that the points are not too close together.
Mathematically, we define the constraint as follows:
\begin{equation}
\forall\mathbf{p}_i, \mathbf{p}_j \in \mathcal{P}, \, \, \, \|\mathbf{p}_i - \mathbf{p}_j\|_2 \geq \delta,
\end{equation}
where \( \delta \) is a pre-defined threshold. However, since the maximization is over a black-box function, we opt for an approximation method to compute the maximization and the constraint part of the system, as detailed in Subsection \ref{sec:3.2}.

\subsubsection{Details of the Algorithm}
\RestyleAlgo{ruled}

\SetKwComment{Comment}{\color{myblue}\# }{}
\begin{algorithm}[hbt!]
\caption{Spatially Adaptive Garment Simulation}\label{alg:two}
$\mathcal{P} \gets \emptyset$\;
$\boldsymbol{\Theta} \gets$ randomly initialized model parameters\;
$N \gets$ the number of sampling points\;
\For{each epoch $e$}{
  \Comment{\color{myblue}Adversarial Player 1}
    $\mu \gets $ user-defined value in $[0, 1]$, with default value $1/2$\;
    $N_a \gets \lfloor \mu N \rfloor$\;
    $N_u \gets N - N_a$\;
    construct and update PDF by \textit{discrete approximation}\;
    $\mathcal{P}_a \gets $ sample $N_a$ points according to the discrete PDF\;
    $\mathcal{P}_u \gets $ uniformly sample $N_u$ points\;
    $\mathcal{P} \gets \mathcal{P}_a \cup \mathcal{P}_u$\;
    $\mathcal{P}^* \gets $ \textit{Lloyd's relaxation} on the $N$ sampling points in $\mathcal{P}$\;
  
  \vspace{2mm}
  \Comment{\color{myblue} Adversarial Player 2}
  \For{$p^* \in \mathcal{P}^*$} {
    $\theta \gets $ random value between $[0, 2\pi/3]$\;
    generate a sampling local structure around $p^*$\;
    rotate the structure using rotation angle $\theta$\;
    compute $\mathcal{L}_{\text{Bend}} (p^*, \boldsymbol{\Theta})$\;
    compute $\mathcal{L}_{\text{Strain}} (p^*, \boldsymbol{\Theta})$\;
    compute $\mathcal{L}_{\text{Gravity}} (p^*, \boldsymbol{\Theta})$\;
    compute $\mathcal{L}_{\text{Collision}} (p^*, \boldsymbol{\Theta})$\;
  }
  $\mathcal{L}_{\text{Sum}} (\mathcal{P}^*, \boldsymbol{\Theta}) \gets 0$\;
  $\mathcal{L}_{\text{Sum}} (\mathcal{P}^*, \boldsymbol{\Theta}) \mathrel{+{=}} \mathcal{W}_{\text{Bend}} \cdot \sum_{p^* \in \mathcal{P}^*} \mathcal{L}_{\text{Bend}}(p^*, \boldsymbol{\Theta}) $\;
  $\mathcal{L}_{\text{Sum}} (\mathcal{P}^*, \boldsymbol{\Theta}) \mathrel{+{=}} \mathcal{W}_{\text{Strain}} \cdot \sum_{p^* \in \mathcal{P}^*} \mathcal{L}_{\text{Strain}}(p^*, \boldsymbol{\Theta})$\;
  $\mathcal{L}_{\text{Sum}} (\mathcal{P}^*, \boldsymbol{\Theta}) \mathrel{+{=}} \mathcal{W}_{\text{Gravity}} \cdot \sum_{p^* \in \mathcal{P}^*} \mathcal{L}_{\text{Gravity}}(p^*, \boldsymbol{\Theta})$\;
  $\mathcal{L}_{\text{Sum}} (\mathcal{P}^*, \boldsymbol{\Theta}) \mathrel{+{=}} \mathcal{W}_{\text{Collision}} \cdot \sum_{p^* \in \mathcal{P}^*} \mathcal{L}_{\text{Collision}}(p^*, \boldsymbol{\Theta})$\;
  $\boldsymbol{\Theta}^* \gets \boldsymbol{\Theta} - \alpha \nabla_{\boldsymbol{\Theta}} \mathcal{L}_{\text{Sum}} (\mathcal{P}^*, \boldsymbol{\Theta})$\;

  $\boldsymbol{\Theta} \gets \boldsymbol{\Theta}^*$
  
}
\Return{$\boldsymbol{\Theta}$}
\end{algorithm}

We provide the pseudocode of the algorithm as in Algorithm \ref{alg:two}. The algorithm uses an innovative adversarial framework that capitalizes on a dual-player system. The goal is to optimize the representation of physical properties of the garment, such as bending, strain, gravity, and collision, while concurrently refining the spatial distribution of the simulation points.

\paragraph{Initialization and Model Setup.}
Initially, the algorithm focuses on setting the groundwork. A set, denoted as \(\mathcal{P}\), is initialized as an empty set which will later serve to store the simulation's sampling points. In parallel, the model parameters, symbolized by \(\mathbf{\Theta}\), are initialized with random values. These could be envisioned as the underlying weights of a MLP or a comparable model, like the parameters in our multi-resolution grid encoding model. These model parameters encodes the neural implicit surface, which represents the shape of the garment. The process is further streamlined by defining \(N\), which represents the total number of desired sampling points in the simulation.

\paragraph{Adversarial Player 1: Optimal Point Sampling.}
In the adversarial training, the first player is responsible for point-sampling to make the sum of losses at these sampling points as large as possible. The parameter, \(\mu\), typically defaults to $0.5$ but remains user-adjustable within the range $[0, 1]$. It divides the total sampling points, \(N\), into two distinct categories: 

\begin{enumerate}
    \item Adaptive Points: A segment of the total points, calculated as \(N_a\), are adaptively sampled. This number is essentially the floor value of the product of \(\mu\) and \(N\).
    \item Uniform Points: The remainder, denoted as \(N_u\), is uniformly sampled. They provide a consistent distribution to make the sampling points have a good coverage of the whole sampling domain.
\end{enumerate}

To achieve the adaptive sampling, a discrete PDF is constructed and updated using the method mentioned in Section ~\ref{sec:3.2}. The $N_a$ adaptive points are then sampled using this discrete PDF, and the $N_u$ uniform points are sampled according to a uniform PDF within the domain. Subsequently, these adaptively and uniformly sampled points are combined into the main set, \(\mathcal{P}\).

To ensure the sampling points do not cluster too closely together, we slightly space out the points but maintain a higher concentration in regions where the losses are large. To achieve this, Lloyd's relaxation is applied. It refines the distribution of the sampling points, ensuring points are spread as uniformly as possible while preserving the original density variations.

\paragraph{Adversarial Player 2: Physical Property Calculation And Optimization.}
As the second adversary enters the game, the focus pivots to the physical essence of the garment. For each point in the optimized set, \(\mathcal{P^*}\), a sampling local structure is generated, as detailed in \ref{sec:3.3}, with a randomly generated rotation angle, \(\theta\), from the range $[0, 2\pi/3]$.

The algorithm then evaluates a suite of loss functions, tailored to measure various physical properties at each point on the neural implicit surface, where the shape of the surface is captured by the current model parameters, \(\mathbf{\Theta}\). This includes determining the garment's bending, strain, gravity, and collision losses. Summing these individual losses across all points, we employ back-propagation in conjunction with gradient descent to refine the model parameters, \(\mathbf{\Theta}\), in order to minimize the loss. This process utilizes a learning rate, \(\alpha\), to update the neural implicit surface.

After iterating between the two adversarial players for the specified epochs, the algorithm concludes, presenting the finely-tuned model parameters, \(\mathbf{\Theta}\). This setup enables continuous querying of the 3D surface positions across the UV domain.

In summary, this novel adversarial framework delivers a more accurate and realistic garment simulation, optimizing spatial representation while capturing the intricate nuances of fabric behavior across diverse situations.

%% file: sec/4_evaluation.tex
\section{Evaluation}
\label{sec:evaluation}
All our tests were conducted on a desktop running Ubuntu 20.04.5 LTS, equipped with an Intel(R) Xeon® E5-1680 v3 @ 3.20GHz processor and a GeForce RTX 2080 Ti graphics card. We developed the framework using Python with Tensorflow 2.0.


\subsection{Network and Encoding}
Given that our training process involves randomly generated sampling points and local structures, ensuring a fair comparison for network and encoding can be challenging if evaluated in an unsupervised scheme. So to achieve our goal of highlighting the efficiency of our proposed multi-resolution grid encoding model, we evaluate various neural network models using a supervised approach. Our ground truth is a manually constructed 3D model that resembles a 3D sine wave. This model incorporates both low and high-frequency details, making it an ideal candidate for assessing the performance of different neural network models.

To ensure a fair comparison, the parameters and sizes of the network models have been fine-tuned so that all models operate under the same memory constraints. 

In detail, the first model is a baseline MLP architecture. It consists of four fully connected layers, each with its weight matrix and bias vector. The input layer has dimensions $2 \times 152$, where 2 represents the dimension of the UV space, followed by two hidden layers with dimensions $152 \times 152$ each, and a final output layer with dimensions $152 \times 3$. This model has a total of $47427$ parameters.

The second model incorporates positional encoding into its architecture. Like the first model, it also consists of four fully connected layers with their respective weight matrices and bias vectors. The input layer has dimensions $18 \times 148$, where $18$ equals the dimension of the UV space plus the hidden dimension of the positional encoding, followed by two hidden layers with dimensions $148 \times 148$ each, and a final output layer with dimensions $148 \times 3$. The total number of parameters in this model is $47363$.

The third model is our multi-resolution grid encoding model, which includes two grid layers with shapes $101 \times 101 \times 3$ and $51 \times 51 \times 3$. These grid layers are followed by four fully connected layers with various weight matrices and bias vectors. The first fully connected layer has dimensions $6 \times 64$, where $6$ represents the number of concatenated grid features, followed by two hidden layers with dimensions $64 \times 64$ each, and a final output layer with dimensions $64 \times 3$, with a total of $47369$ parameters.

\paragraph{Speed Comparison.}
 A comparative analysis of the running times across various neural network models is provided in Table ~\ref{tab:model_running_times}. In terms of the number of epochs required for convergence, the baseline MLP model was trained for $500000$ epochs before completion, while the multi-resolution grid encoding model was trained for only $400$ epochs. In terms of clock running time, the multi-resolution grid encoding model is approximately $346.21$ times faster than the Baseline MLP model. The baseline MLP model took $2$ hours, $41$ minutes, and $34$ seconds to reach its final epoch, while the multi-resolution grid encoding model only took $28$ seconds.
 \begin{table}[h]
    \centering
    \small
    \begin{tabular}{@{}lrrr@{}}
        \toprule
        Network Model & End Epoch & Running Time \\
        \midrule
        Baseline MLP & 500000 & 2h 41m 34s \\
        Positional Encoding & 56000 & 23m 1s \\
        Multigrid Encoding & \textbf{400} & \textbf{28s} \\
        \bottomrule
    \end{tabular}
    \vspace{2mm}
    \centering
    \caption{Running Time Comparison for Different Network Models}
    \label{tab:model_running_times}
\end{table}
 
\begin{figure*}[tbp]
  \centering
  \mbox{} \hfill
  \includegraphics[width=1\linewidth]{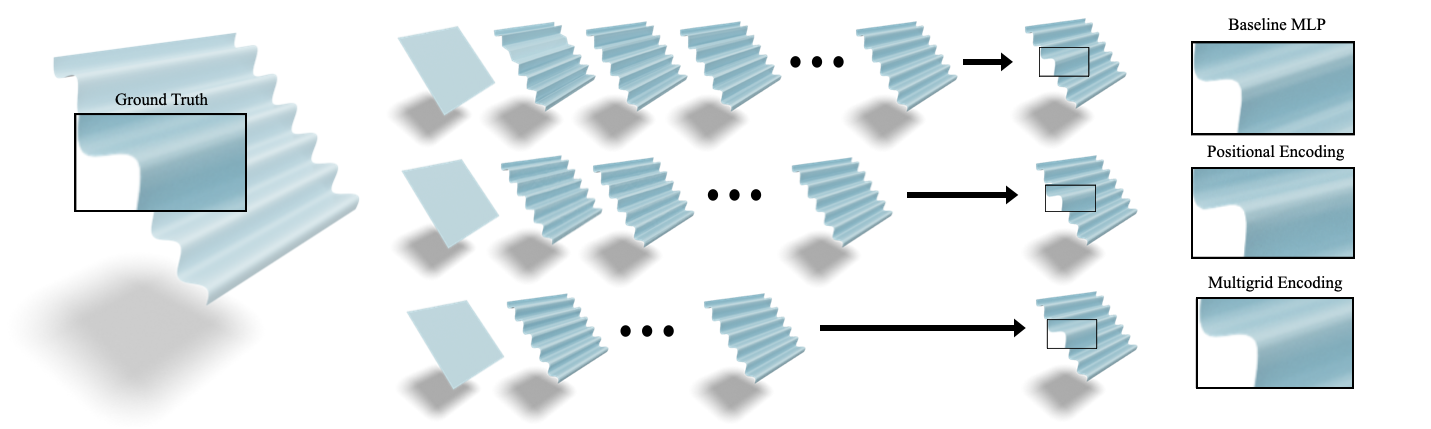}
  \caption{\label{fig:encoding_zoomed}%
          \textbf{The Power of Multi-Resolution Grid Encoding.} In supervised training, the use of multi-resolution grid encoding results in training speeds that are $346.21$ times faster compared to the baseline MLP model, $49.32$ times faster compared to the positional encoding model. Additionally, it facilitates the capture of high-frequency details in the ground truth more effectively compared to both the baseline MLP model and the positional encoding model.}
\end{figure*}

\paragraph{Quality Comparison.} For a direct visual comparison of the fully trained outputs, please refer to Figure \ref{fig:encoding_zoomed}. It is worth noting that both the baseline MLP model and the positional encoding model exhibit some minor artifacts in their outputs despite trained for much longer time, which you may zoom in to see clearly. These artifacts manifest as challenges in maintaining sharp and high-frequency features. In contrast, the multi-resolution grid encoding model produces results that closely align with the ground truth, showcasing a high level of fidelity in its representation.

\begin{figure*}[tbp]
  \centering
  \mbox{} \hfill
  \includegraphics[width=1\linewidth]{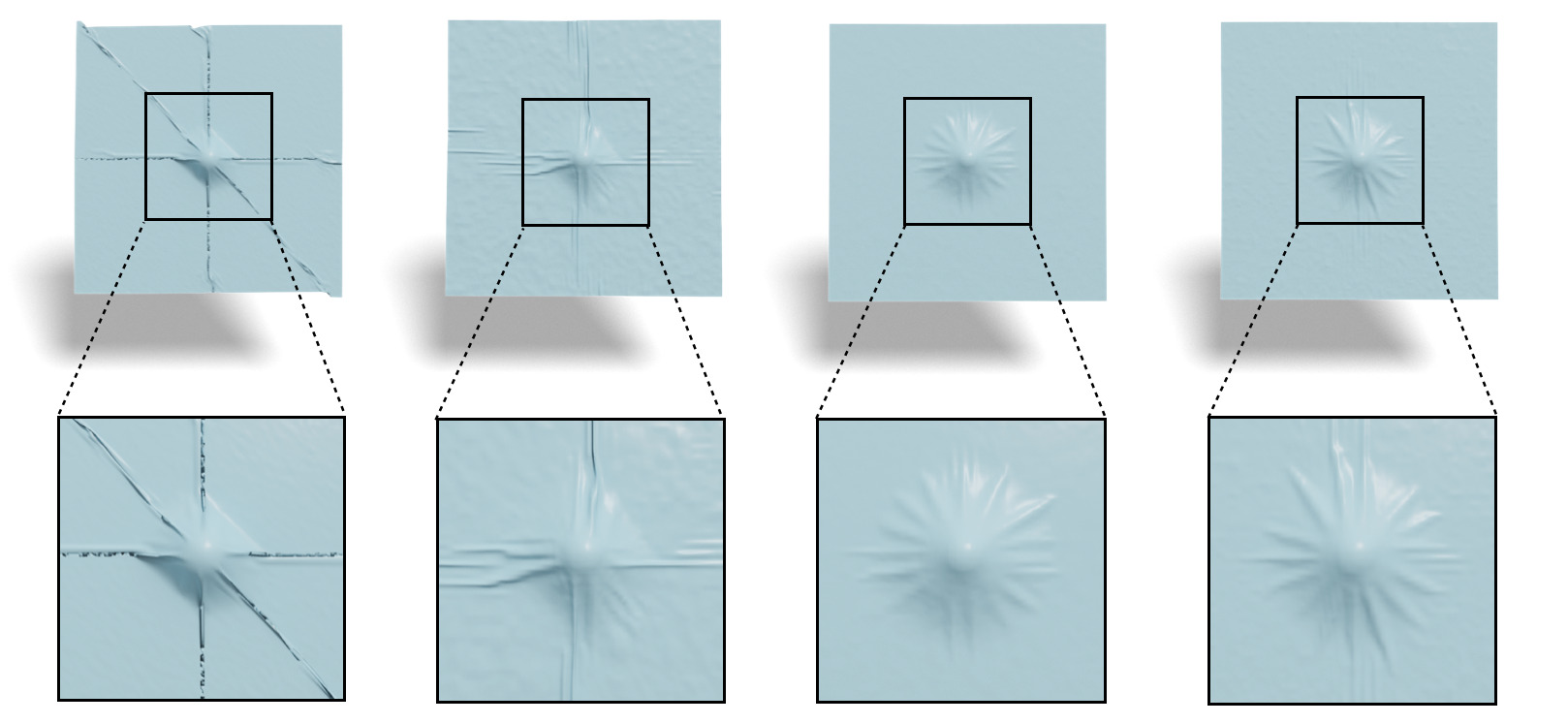}
  \caption{\label{fig:ball_zoomed}%
\textbf{Wrinkle Generation in Cloth-Ball Interaction.} 
(Left) A coarse-resolution mesh grid (resolution \(128 \times 128\), totaling \(49152\) free variables) employs the original mesh connectivity for loss computation. The wrinkles are generated in certain directions only, leading to severe artifacts due to discretization. 
(Middle Left) A multi-resolution grid neural network with fewer free variables (\(47369\)) captures cloth details using the original mesh connectivity. The wrinkles are still generated in certain directions only, resulting in fewer artifacts but making convergence challenging. 
(Middle Right) The same variables (\(47369\)) in the multi-resolution grid model, with losses computed using our novel method and \textit{uniform} sampling of \textit{local structures}. When converged, it provides wrinkles in all directions and almost no artifacts. 
(Right) The same variables (\(47369\)) in the model, with losses computed using our method and \textit{adaptive} sampling of \textit{local structures}, resulting in similar or even more enhanced wrinkles with fewer epochs needed for training compared to uniform sampling.}

\end{figure*}

\begin{figure*}[tbp]
  \centering
  \mbox{} \hfill
  \includegraphics[width=1\linewidth]{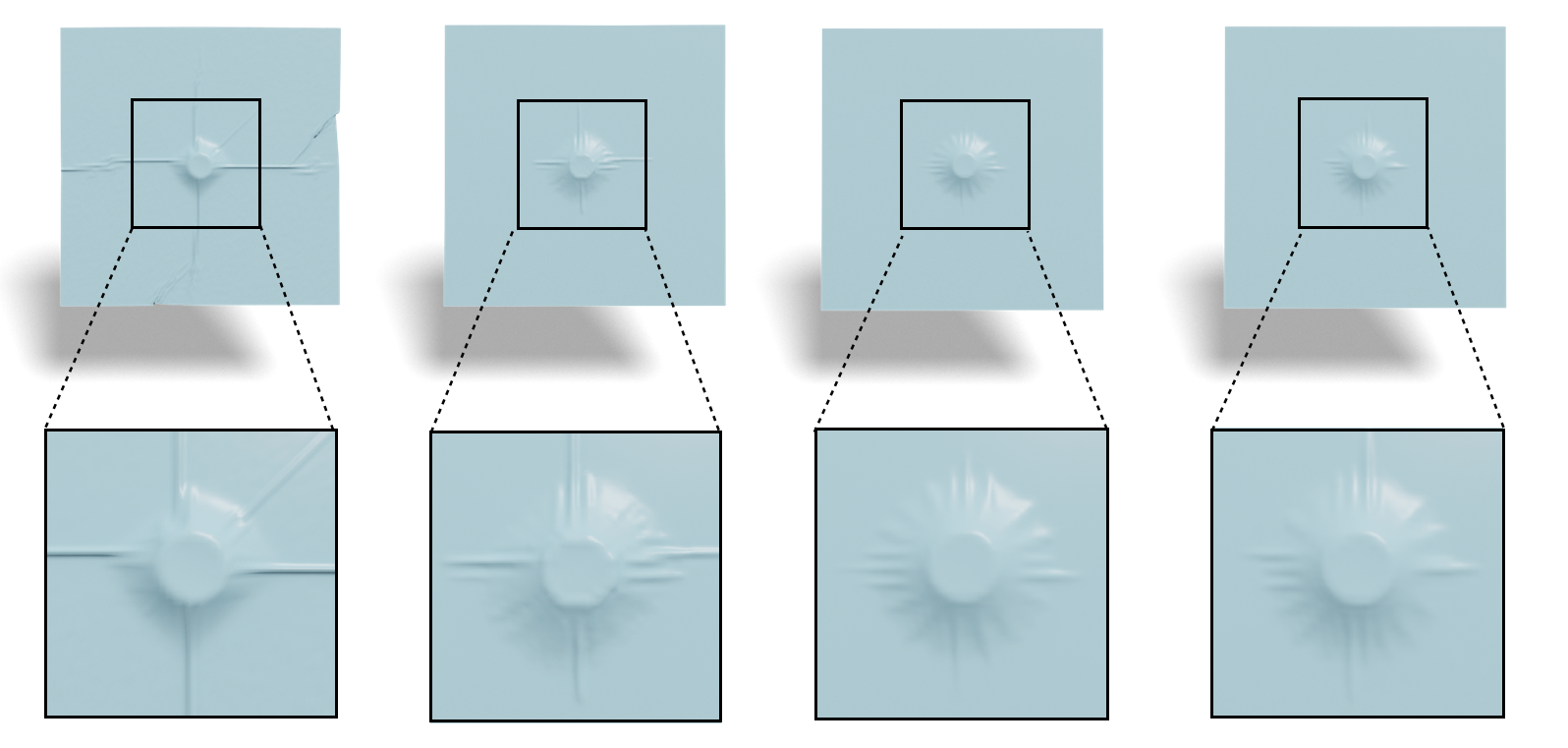}
\caption{\label{fig:torus_zoomed}%
\textbf{Wrinkle Generation in Cloth-Torus Interaction.} 
(Left) A coarse-resolution mesh grid (resolution \(128 \times 128\), totaling \(49152\) free variables) employs the original mesh connectivity for loss computation. Due to the discrete low-resolution mesh structure, wrinkles appear only in limited directions, with unnatural artifacts. 
(Middle Left) A multi-resolution grid neural network with fewer free variables (\(47369\)) captures cloth details using the original mesh connectivity. The wrinkle pattern is improved; however, it is still challenging to show detailed wrinkles, and the resulting pattern is not symmetric. 
(Middle Right) The same variables (\(47369\)) in the multi-resolution grid model, with losses computed using our novel method but with \textit{uniform} sampling of \textit{local structures}. The wrinkle pattern is significantly improved with natural details. 
(Right) The same variables (\(47369\)) in the model, with losses computed using our method and \textit{adaptive} sampling of \textit{local structures}, yield the most natural and refined wrinkles and require fewer training epochs to converge when compared to uniform sampling.}

\end{figure*}

\begin{figure*}[tbp]
  \centering
  \mbox{} \hfill
  \includegraphics[width=1\linewidth]{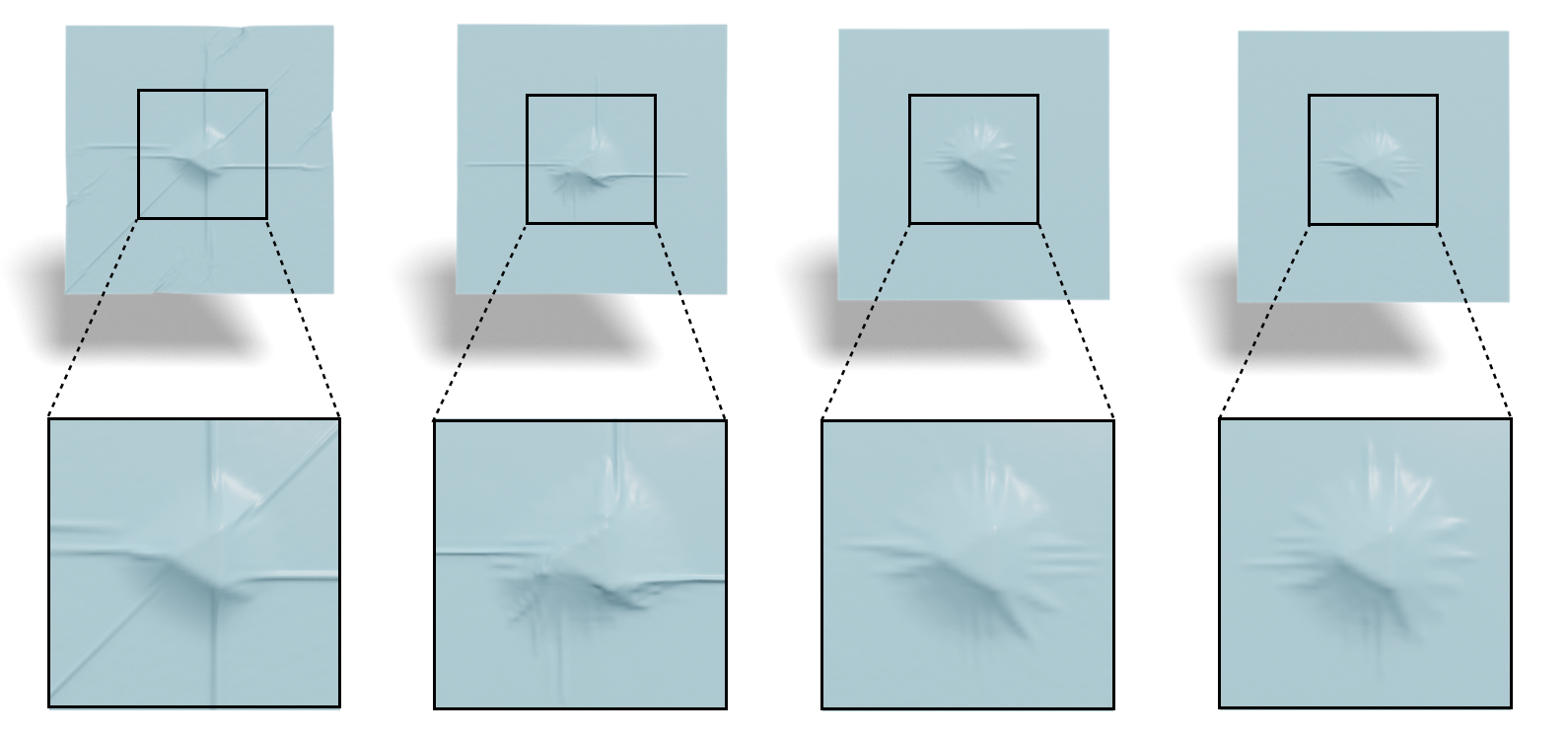}
\caption{\label{fig:triangle_zoomed}%
\textbf{Wrinkle Generation in Cloth-Prism Interaction.} (Left) A coarse-resolution mesh grid (resolution \(128 \times 128\), totaling \(49152\) free variables) employs the original mesh connectivity for loss computation. The wrinkle pattern contains significant artifacts and appears only in random directions. (Middle Left) A multi-resolution grid neural network with fewer free variables (\(47369\)) captures cloth details using the original mesh connectivity. It is difficult for the model to converge, resulting in unnatural wrinkles. (Middle Right) The same variables (\(47369\)) in the multi-resolution grid model, with losses computed using our novel method but with \textit{uniform} sampling of \textit{local structures}, learn a symmetric wrinkle pattern, representing a significant improvement in the results. (Right) The same variables (\(47369\)) in the model, with losses computed using our method and \textit{adaptive} sampling of \textit{local structures}, yield even more enhanced wrinkles with significantly fewer epochs needed for training.}

\end{figure*}

\subsection{Representation}
We compare the implicit neural representation using the multi-resolution grid encoding model with traditional mesh representations, employing the same unsupervised losses computed based on the original mesh connectivity.

When employing the traditional mesh representation, the simulation simplifies to a vertex optimization problem, with the free variables set to the 3D positions of the vertices (for a total of $49152$ free variables in our settings). These variables are optimized to minimize the weighted sum of the losses calculated using the mesh connectivity. 

On the other hand, when utilizing the implicit neural representation of the surface with our multi-resolution grid encoding model, the free variables are the parameters within the network model, totaling $47369$ free variables, which is fewer than in vertex optimization using the traditional mesh representation. In this scenario, since the input to the network model can be any 2D UV point, we can compute the 3D real-world position of the corresponding UV point using the network's output. Thus, we can query the 3D deformed positions of the original input mesh vertices (in UV space) and compute the losses using the original mesh connectivity. The network parameters will then be optimized to minimize the weighted sum of these losses.

\begin{figure}[tbp]
  \centering
  \mbox{} \hfill
  \includegraphics[width=1\linewidth]{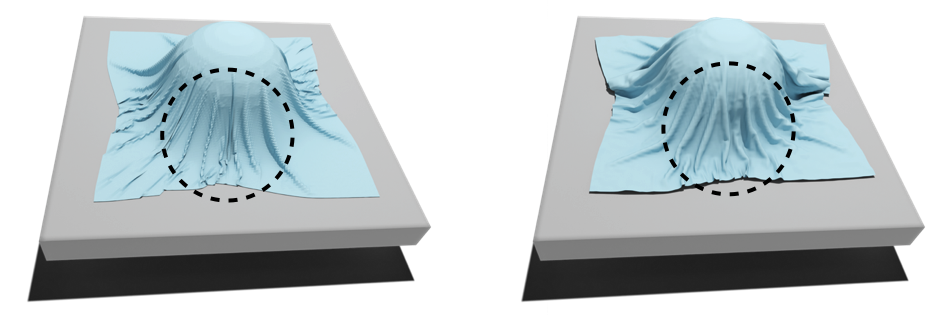}
  \caption{\label{fig:sphere_zoomed}%
           \textbf{In Comparison to Traditional Mesh Representation.} With fewer free variables, the implicit neural representation on the right was able to produce more natural wrinkles and contained far fewer artifacts compared to the traditional mesh representation at low resolution.}
\end{figure}

\paragraph{Quality Comparison.} For a direct visual comparison, please refer to the first two columns in Figure ~\ref{fig:ball_zoomed}, Figure ~\ref{fig:torus_zoomed}, and Figure ~\ref{fig:triangle_zoomed}. These examples were intentionally created to highlight the capability of a single source of cloth-object interaction in generating predictable localized wrinkles. Additionally, you can examine the first two columns of Figure ~\ref{fig:teaser}, and Figure ~\ref{fig:sphere_zoomed} for more complicated examples and an overall effect. 

In all these examples, we observe that when the traditional mesh representation is used, the expressiveness of the local wrinkles is restricted by the discretization. Moreover, when the mesh resolution is low, the local wrinkles are either ignored or become artifacts.

In contrast, when utilizing the implicit neural representation within our multi-resolution grid encoding model, we observe fewer artifacts. Nevertheless, it is crucial to emphasize that this comparison primarily delves into exploring the representation aspect, with losses computed based on the original mesh connectivity. The extent of improvement may not be as pronounced at this stage, as the implicit neural representation provides the benefits of a continuous domain and adaptivity, enabling us to compute local losses without the limitations imposed by discretization. We will further showcase this capability in the next subsection.

\begin{figure}[tbp]
  \centering
  \mbox{} \hfill
  \includegraphics[width=1\linewidth]{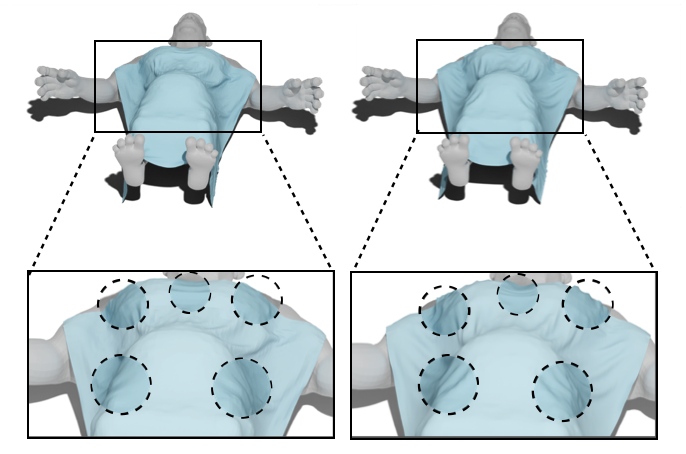}
\caption{\label{fig:ogre_zoomed}%
  \textbf{Effectiveness of Adaptive Sampling (Ogre Test).} Utilizing the same multi-resolution encoding neural network model and conducting experiments over the same number of epochs, the adaptive sampling approach on the right remarkably enhanced the wrinkle patterns in detailed regions compared to the uniform sampling approach on the left.}
\end{figure}

\begin{figure}[tbp]
  \centering
  \mbox{} \hfill
  \includegraphics[width=1\linewidth]{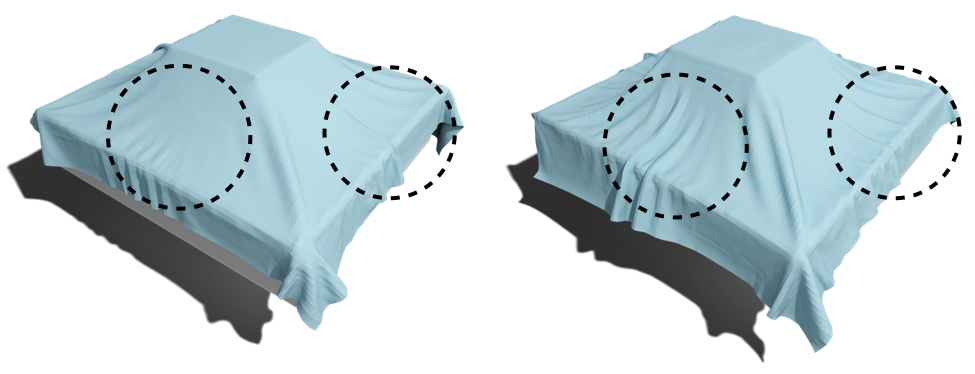}
  \caption{\label{fig:table_zoomed}%
           \textbf{Effectiveness of Adaptive Sampling (Table Test).}When using an identical number of epochs and the same multi-resolution encoding neural network model, the adaptive sampling method demonstrated superior capability in capturing intricate and improved wrinkle patterns compared to the uniform sampling approach.}
\end{figure}

\subsection{Simulation Losses}
To better showcase the superiority of our novel loss computation method on top of the neural implicit surface, we compare the simulation results achieved using different unsupervised loss computation methods when using the same network and encoding architecture.
\paragraph{Quality Comparison.}
For a direct visual comparison, please refer to the second and third columns in Figure ~\ref{fig:ball_zoomed}, Figure ~\ref{fig:torus_zoomed}, and Figure ~\ref{fig:triangle_zoomed} to closely examine local and detailed wrinkles. Furthermore, you can explore the second and third columns of Figure ~\ref{fig:teaser} for more complex examples and an overall view.

In all of these examples, we have observed that when the simulation losses are determined based on the original mesh connectivity, the neural network parameters tend to capture localized wrinkles less effectively. This is primarily due to the limited utilization of the continuous domain; we consistently query the same UV points and train the network on these discrete points, causing the network parameters to be updated solely based on results computed at these specific points within the continuous domain.

However, when the simulation losses are computed using our innovative approach of sampling local structures, the continuous domain can be thoroughly explored. In each epoch, we query random sampling points within the continuous domain and optimize the network parameters based on the losses computed at these points. This results in significant improvements, particularly noticeable in the case of localized wrinkles generated by a single source of cloth-object interaction. The wrinkles are much better captured when the losses are computed based on our novel local structure sampling method.

\subsection{Adaptivity}
\paragraph{Speed Comparison.} We kept all other settings the same while changing only the sampling method. We compared the uniform sampling method to the adaptive sampling method, and it turned out that the adaptive sampling method resulted in faster convergence. We compare the number of epochs required for convergence and present the results in Table ~\ref{tab:num_epochs_convergence}.
 \begin{table}[h]
    \centering
    \small
    \begin{tabular}{@{}lrrr@{}}
        \toprule
        Model Name & Figure & Uniform Sampling & Adaptive Sampling \\
        \midrule
        Ball & \ref{fig:ball_zoomed} & 14900 & 11950 \\
        Torus & \ref{fig:torus_zoomed}& 20500 & 17650 \\
        Prism & \ref{fig:triangle_zoomed} & 13950 & 8450 \\
        \bottomrule
    \end{tabular}
    \vspace{2mm}
    \centering
    \caption{Number of Epochs Required for Convergence}
    \label{tab:num_epochs_convergence}
\end{table}

\paragraph{Quality Comparison.} We keep the number of epochs constant, as summarized in Table ~\ref{tab:same_num_epochs}, and visually compare the simulation results of different models. For a direct visual comparison, please refer to the last two columns in Figure ~\ref{fig:teaser}, Figure ~\ref{fig:ogre_zoomed}, and Figure ~\ref{fig:table_zoomed} to closely examine the detailed wrinkles.
 \begin{table}[h]
    \centering
    \small
    \begin{tabular}{@{}lrrr@{}}
        \toprule
        Model Name & Figure & Uniform Sampling & Adaptive Sampling \\
        \midrule
        Scorpion & \ref{fig:teaser} &  2400 & 2400\\
        Ogre & \ref{fig:ogre_zoomed} &  3450 & 3450 \\
        Table & \ref{fig:table_zoomed} &  1850 & 1850\\
        \bottomrule
    \end{tabular}
    \vspace{2mm}
    \centering
    \caption{Number of Epochs Used for Training}
    \label{tab:same_num_epochs}
\end{table}

When trained with the same number of epochs, the simulation results were significantly improved when using adaptive sampling. This is because the system placed greater emphasis on regions requiring more attention during adaptive sampling, leading to deeper and clearer wrinkles in the results.

\begin{table*}[h]
    \centering
    \small
    \begin{tabular}{@{}lrrrrrrrrrrrrrrr@{}}
        \toprule
        Model Name & Figure & $\alpha$ & $w_\text{Strain}$ & $w_\text{Bend}$ & $w_\text{Gravity}$ & $w_\text{Collision}$ &  Cloth Resolution & Body Vertex Count & Triangle Side Length\\
        \midrule
        Scorpion & \ref{fig:teaser} & 0.0005 & 0.005 & 0.0005 & 2 & $10^7$ & $128 \times 128$ & 49997 & 0.001\\ 
        Ball & \ref{fig:ball_zoomed} & 0.0005 & 0.005 & 0.0005 & 2 & $10^7$ & $128 \times 128$ & 25060 & 0.001\\
        Torus & \ref{fig:torus_zoomed} & 0.0005 & 0.005 & 0.0005 & 2 & $10^6$ & $128 \times 128$ & 25154 & 0.001\\
        Prism & \ref{fig:triangle_zoomed} & 0.0005 & 0.005 & 0.0005 & 2 & $10^6$ & $128 \times 128$ & 40964 
 & 0.001\\
        Ogre & \ref{fig:ogre_zoomed} & 0.0005 & 0.005 & 0.0005 & 2 & $10^7$ & $128 \times 128$ & 62194 & 0.001\\
        Table & \ref{fig:table_zoomed} & 0.0005 & 0.005 & 0.0005 & 2 & $10^7$ & $128 \times 128$ & 40964 & 0.001\\
        Sphere & \ref{fig:sphere_zoomed} & 0.0005 & 0.005 & 0.0005 & 2 & $10^7$ & $128 \times 128$ & 25060 & 0.001\\
        \bottomrule
    \end{tabular}
    \vspace{2mm}
    \centering
    \caption{Parameter Settings for the Models Used for Experiments}
    \label{tab:same_num_epochs}
\end{table*}

%% file: sec/5_limitations_and_conclusion.tex
\section{Limitations and Conclusion}
\label{sec:limitations and conclusion}

In this paper, we delved into the potential of leveraging implicit neural representations to simulate intricate cloth details, such as wrinkles. Through various cloth-object interaction examples, our technique demonstrates superiority over conventional discrete representations under the same memory constraints. This is most evident in the enhanced simulation of detailed cloth wrinkles, especially the fine and localized ones. However, our work does come with its challenges. We have categorized these into five aspects, summarized as follows:

\paragraph{UV Mapping Limitation.} Our current model is restricted to a straightforward case where the UV space is a square domain, $[0, 1]^2$. When extending this to complex garments, the UV map might encompass irregular boundaries, seams, and void regions. One approach to address this is segmenting the UV map into panels and using a mask within each panel to highlight void areas. Deformations are then learned only for UV positions outside these void spaces. For managing seams and boundaries, constraints could be introduced to ensure smooth transitions on either side of the seams. While we currently adjust the sampling local structures to fit the square domain, future research could delve into improved methods, possibly exploring boundary-specific sampling structures or mirrored seam padding.

\paragraph{Theoretical Guarantee.} While our adaptive method has demonstrated promising experimental outcomes, a rigorous proof might be necessary to provide a solid theoretical foundation. This involves proving that such adaptive sampling would closely approximate the ideal optimization scenario, aiming to minimize the integrated losses across the entire domain over an infinite number of sampling points.

\paragraph{Sampling Methods.} We have examined and compared three different sampling techniques: discrete PDF approximation, as well as other probabilistic techniques such as simulated annealing and Bayesian optimization with Gaussian processes. Among these three, discrete PDF approximation performs the best in our specific settings. However, there are numerous other methods that could be applicable. For example, the Winner-takes-it-all method involves multiple random samplings and selecting the one yielding the maximum sum of function values. While this method might appear time-intensive, its practicality might be feasible considering the amortized runtime.

\paragraph{Encoding Models.} Our neural network currently employs a multi-resolution grid encoding, which considerably accelerates the process compared to the baseline MLP. Numerous encoding models exist in other related research field, such as multi-resolution hash encoding. Incorporating hash encoding alongside our grid encoding is a potential avenue for enhancement, though its efficacy remains contingent on the specific problem.

\paragraph{Loss Balance.} The weights for the losses in our system are adjusted manually for every model. A more sophisticated method to determine the loss balance based on material attributes could make tuning more straightforward. However, given the geometrical nature of the collision loss and the unpredictability of the sampling process, devising a systematic method for determining loss weights might prove challenging.\\

In conclusion, our methodology exhibits promising results in simulating cloth details. However, there is room for continued research and enhancement. We look forward to seeing subsequent studies refine and expand upon our approach. Particularly when considering the simulation of characters in tight-fitting clothing with wrinkles arising from garment-body collisions, the potential is vast. This opens doors for innovative applications in sectors like fashion design, virtual try-ons, and animation.